# Inferring High-level Geographical Concepts via Knowledge Graph and Multi-scale Data Integration: A Case Study of C-shaped Building Pattern Recognition

Zhiwei Wei[a,b], Yi Xiao[c,d], Wenjia Xu[e], Mi Shu[f], Lu Cheng[c],Yang Wang[a,b], Chunbo Liu[a,b]

[a]*Key Laboratory of Network Information System Technology (NIST), Aerospace Information Research Institute, Chinese Academy of Sciences, Beijing 100190, China*; [b]*The Aerospace Information Research Institute, Chinese Academic of Sciences, Beijing 100190, China*; [c]*School of Resource and Environment Science, Wuhan University, Wuhan 430079, China*; [d]*School of Software Engineering, Shenzhen Institute of Information Technology, Shenzhen 518172, China*; [e]*School of Information and Communication Engineering, Beijing University of Posts and Telecommunications, Beijing 100876, China*; [f]*Institute of Remote Sensing and GIS, Peking University, Beijing 100871, China.*

# Inferring High-level Geographical Concepts via Knowledge Graph and Multi-scale Data Integration: A Case Study of C-shaped Building Pattern Recognition

**Abstract**: Effective building pattern recognition is critical for understanding urban form, automating map generalization, and visualizing 3D city models. Most existing studies use object-independent methods based on visual perception rules and proximity graph models to extract patterns. However, because human vision is a part-based system, pattern recognition may require decomposing shapes into parts or grouping them into clusters. Existing methods may not recognize all visually aware patterns, and the proximity graph model can be inefficient. To improve efficiency and effectiveness, we integrate multi-scale data using a knowledge graph, focusing on the recognition of C-shaped building patterns. First, we use a property graph to represent the relationships between buildings within and across different scales involved in C-shaped building pattern recognition. Next, we store this knowledge graph in a graph database and convert the rules for C-shaped pattern recognition and enrichment into query conditions. Finally, we recognize and enrich C-shaped building patterns using rule-based reasoning in the built knowledge graph. We verify the effectiveness of our method using multi-scale data with three levels of detail (LODs) collected from the Gaode Map. Our results show that our method achieves a higher recall rate of 26.4% for LOD1, 20.0% for LOD2, and 9.1% for LOD3 compared to existing approaches. We also achieve recognition efficiency improvements of 0.91, 1.37, and 9.35 times, respectively.

Keywords: building; pattern recognition; knowledge graph; rule-based reasoning; Gestalt principals.

## 1. Introduction

Geographical concepts at a high level are essential in revealing the specific semantics of spatial objects, and their automatic inference can significantly enhance users' understanding and application of spatial data (Wei et al., 2018). For instance, buildings may exhibit high-level patterns such as alphabetical shapes or linear arrangements due to underlying socio-political, economic, cultural, or natural factors, and identifying these patterns plays a vital role in urban understanding (Du et al., 2019; Yan et al., 2019), 3D city model visualization (Mao et al., 2012; Hu et al., 2018), and automatic map generalization (Renard et al., 2014; Shen et al., 2022).

Various methods have been proposed to automatically recognize these patterns, which can be classified into individual-level shape recognition and group-level arrangement recognition. Shape recognition methods represent building shapes using turning functions (Arkin et al., 1999), shape contexts (Belongie et al., 2002), Fourier descriptors (Ai et al., 2013), graph convolutional autoencoder models (Yan et al., 2020), and rectangular encoding (Wei et al., 2021), and then classify building shapes into cognitively enhanced shapes, like C-shaped or E-shaped ones. Arrangement recognition methods involve recognizing regularly arranged building clusters, such as alphabetical or linear ones, using various techniques, including template matching (Rainsford & Mackaness, 2002; Xing et al., 2021), structural rules (Zhang et al., 2013; Du et al., 2015; Pilehforooshha & Karimi, 2018; Wang & Burghardt, 2020), or machine learning approaches (He et al., 2018; Yan et al., 2019; Zhao et al., 2020). In some cases, arrangement recognition is also referred to as pattern recognition. To avoid confusion, in this work, pattern recognition will only refer to both shape recognition and arrangement recognition, while arrangement recognition solely refers to recognizing regularly arranged building clusters.

Although various approaches have been proposed to recognize different patterns, most existing studies rely on object-independent methods to extract patterns based on proximity graph models and present two significant challenges. First, object-independent pattern recognition may not identify all visually apparent patterns. For instance, shape recognition methods can only detect the C-shaped building in Figure 1(d), while arrangement recognition methods can only detect the C-shaped pattern in Figure 1(b). However, the buildings in Figures 1(a) and 1(c) are also visually C-shaped but cannot be recognized by the above methods. Research in psychology has also shown that human vision is a part-based system, which means shapes sometimes need to be decomposed into parts or grouped into clusters for visual tasks (Singh et al., 1999). Second, arrangement recognition can be time-consuming as it involves many searching and matching operations due to the nature of proximity graph models. This low-efficiency bottleneck limits the application of such methods to large areas.

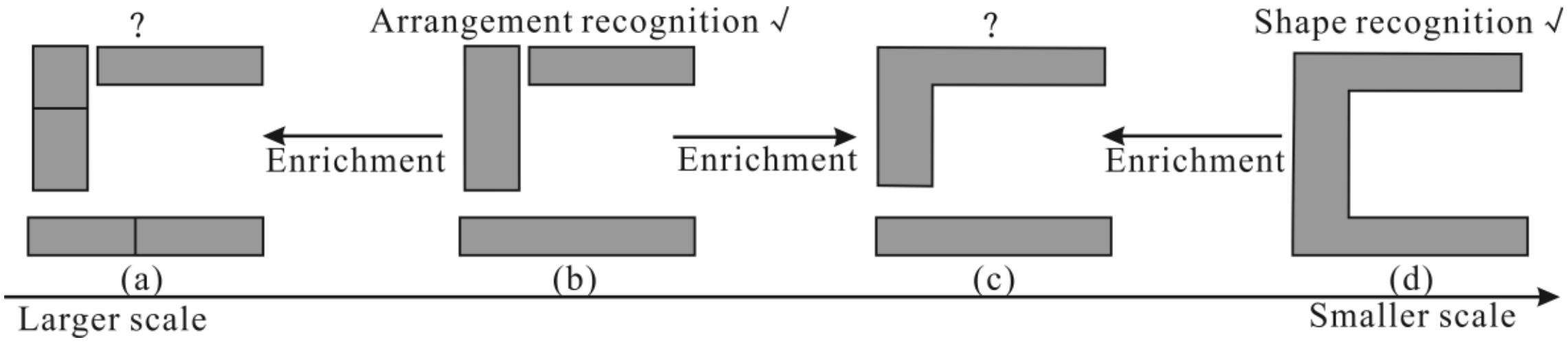

Figure 1. An example to illustrate the limitations of existing pattern recognition methods and the potential benefits of integrating multi-scale data. (a), (c) The patterns cannot be recognized using existing methods. (b) The pattern can be recognized using arrangement recognition; (d) The pattern can be recognized using shape recognition.

Fortunately, over the past few decades, there has been a tremendous increase in the production of geospatial data at different scales (Zhang et al., 2018). This data reflects user understanding of how to cluster or divide buildings at varying levels. By integrating multi-scale data, it is possible to recognize visually distinct patterns more effectively. For instance, existing methods can successfully recognize the building patterns in Figures 1(b) and 1(d). If the corresponding relations between buildings are identified across Figures 1(a)-1(d), an enrichment process based on the results from Figures 1(b) and 1(d) may facilitate the recognition of the patterns in Figures 1(a) and 1(c). Moreover, emerging Semantic Web technologies provide useful tools for data integration, with the knowledge graph-based method receiving considerable attention due to its extensive applications and infrastructure, such as various graph databases (Huang et al., 2020). Since entities and their relationships in a knowledge graph are stored as key-value pairs in a graph database, there is no need to consume computing resources to establish matching between entities, thus enabling efficient reasoning for new knowledge. Consequently, representing the properties and relationships of buildings within and across different scales using a knowledge graph may significantly improve the efficiency and accuracy of building pattern recognition.

Motivated by the ideas mentioned above, we aim to enhance the efficiency and effectiveness of building pattern recognition by integrating multi-scale data through a knowledge graph. In this study, we have selected C-shaped building patterns as a case study, which can serve as a representative example for recognizing other patterns as well. Our method involves constructing a specialized knowledge graph that captures the relationships between buildings at various scales and enables us to recognize C-shaped patterns more accurately. We store this knowledge graph in a graph database and convert the rules for C-shaped pattern recognition and enrichment into query conditions on the knowledge graph. By executing graph queries on the knowledge graph, we can identify and enrich the C-shaped patterns efficiently.

## 2. Related works

### 2.1 *Building shape recognition*

Building shape recognition is crucial and relies heavily on effective shape representation. There are two primary methods for shape representation: region representation and boundary representation. Region representation involves measuring the overall characteristics of the building’s region, such as compactness (Li et al., 2013) or, the equal area circle (Basaraner and Cetinkaya 2017). The European Agent program also conducted a detailed analysis of the measures for building characteristics. Among them, Duchêne et al. (2003) studied the measures for building orientation, while Burghardt et al. (2005) studied the principal components of building spatial features. Wei et al. (2018) analyzed in detail 24 measures for describing building geometric features and proposed a measure set for building geometry analysis. Boundary representation approximates the shape of a string or function, such as turning functions (Arkin et al., 1999), shape contexts (Belongie et al., 2002), Fourier descriptors (Ai et al., 2013), and rectangular encoding (Wei et al., 2022). Additionally, artificial intelligence algorithms such as graph convolutional autoencoder models or CNN-based polygon encoders have also been introduced for shape representation (Yan et al., 2020; Mai et al., 2022). Using these shape representation methods, building shapes can then be classified into standard shapes, such as C-shaped, E-shaped, F-shaped, I-shaped, and others. These shapes are commonly used in building typification and 3D city model visualization (Yan et al., 2016; Xu et al., 2018).

In summary, existing approaches have been able to effectively recognize the standard shapes. However, these shapes represent only a part of building patterns. To achieve a more comprehensive method of pattern recognition, it is necessary to consider other information, such as patterns recognized using arrangement recognition at the same scale or patterns recognized at other scales.

## 2.2 *Building arrangement recognition*

Building arrangement patterns refer to buildings arranged regularly, and the key to recognizing such patterns is to define their regular arrangements. This can be achieved through template-based, structural rule-based, or machine learning-based methods.

Template-based methods formalize building patterns as predefined templates. For instance, Rainsford and Mackaness (2002) used this idea to detect linear patterns in rural buildings, while Xing et al. (2021) employed templates to detect combined collinear patterns. Similarly, Yang (2011) and Gong et al. (2018) introduced stair-shaped, Z-shaped, and H-shaped templates defined by structural parameters. Nonetheless, these methods have limitations in recognizing all potential patterns due to their rigid formalization. Alternatively, machine learning-based methods employ intelligent algorithms such as graph convolutional networks and random forest algorithms to learn the rules for pattern recognition (He et al., 2018; Yan et al., 2019; Zhao et al., 2020). However, these methods may require a large training dataset from diverse regions to enhance their accuracy. Additionally, an examination of the reasons why these algorithms are effective may also be necessary.

Structural rule-based methods define patterns as groups of buildings based on perceptual rules. Compared to the aforementioned methods, structural rule-based methods provide greater flexibility in defining patterns, making them widely used in current research. They primarily utilize the proximity graph model to represent the properties and relationships of buildings, and rules are subsequently defined to detect patterns. For example, Zhang et al. (2013) recognized collinear and curvilinear patterns based on the minimum spanning tree (MST) with Gestalt principles. Similarly, Wei et al. (2018) detected linear patterns using four proximity graphs, where the relative nearest graph (RNG) proved to be the most accurate. Pilehforooshha and Karimi (2018) proposed a framework for detecting linear building patterns based on a defined similarity index, which was then refined by a pattern interaction index. Wang and Burghardt (2020) employed stroke principles to detect linear building patterns based on a proximity graph. As different linear building patterns may share the same buildings, Gong et al. (2014) detected multi-connected linear patterns by deleting edges from a DT-like proximity graph with structural parameters, grid patterns are then recognized based on these multi-connected linear patterns. As human vision is known to be a part-based system, Du et al. (2015) developed a three-level relational method based on spatial reasoning to detect collinear patterns and alphabetical-shaped patterns. In this method, qualitative angle and qualitative size relations between buildings were defined to link them to building patterns. In another study, Wei et al. (2022) recognized linear building patterns combining convex polygon decomposition, where a building can be decomposed into sub-buildings for pattern recognition. For 3D city model visualization, Mao et al. (2012) detected linear patterns on the ground plan of a building model using the MST. However, many searching and matching operations are required for these methods to detect patterns based on proximity graph models, making them inefficient.

To summarize, structural rule-based approaches are popular in current research due to their flexibility in meeting personalized user needs. However, their efficiency is limited by the graph traversal strategy, which hinders their application in large-scale regions. Therefore, it is imperative to develop more efficient methods. Moreover, recent studies have demonstrated that incorporating data at lower or higher levels can enhance building pattern recognition. For instance, Wei et al. (2022) showed that more linear patterns can be detected by decomposing a building into sub-buildings, while Xing et al. (2021) demonstrated that grouping buildings into a cluster can enable the recognition of more collinear patterns. Therefore, integrating multi-scale data could improve the effectiveness of pattern recognition.

### 2.3 *Knowledge graph representation and rule-based reasoning*

The property graph model is widely used for representing knowledge graphs, and in this approach, we use it to replace the proximity graph for building pattern recognition (Wang et al., 2019). To store the resulting knowledge graph, we utilize Neo4j, a popular graph database. The Cypher query language in Neo4j is then applied to express the rules governing building pattern recognition. Consequently, in this section, we introduce the utilization of the property graph model in Neo4j for knowledge graph representation and rule-based reasoning. It should be noted that other graph models such as the resource description framework (RDF) and graph database could also be used for this purpose.

A property graph in Neo4j can be represented as $G=(V,E)$, $V=\{v_1,v_2,...,v_n\}$ is an entity set, $E=\{e(v_i,v_j),v_i\in V,v_j\in V\}$ is a relationship set. Each entity, denoted by $v_m$, may have zero or more labels that indicate its type, and each relationship between entities $v_i$ and $v_j$, denoted by $e(v_i,v_j)$, must have a type to define what type of relationship they are. Both entities and relationships have properties (key-value pairs), which further describe them. In Figure 2, the right part illustrates demonstrates how a property graph can represent the proximity relations (Has_Proxi), area similarity relations (Has_Sim), and area similarity degree (SimDeg) among three buildings. Compared to a representation using a proximity graph (the left part of Figure 2), $e(v_i,v_j)$ in a property graph can represent various relations, while the $e(v_i,v_j)$ in a proximity graph can only represent proximity relations. It is worth noting that the proximity graph and the property graph can have different expressions according to needs in practice, Figure 2 is presented solely to provide examples.

Once a knowledge graph has been constructed and stored, new knowledge can be detected by implementing rules as query conditions using the Cypher language. For example, to identify buildings that have exactly two neighbors, the rules can be defined as presented in Table 1. By translating these rules into Cypher query conditions, we can retrieve buildings that satisfy the specified criteria. In this case, the building with bID 1 can be obtained.

Table 1. Rules via Cypher language to identify buildings that have exactly two neighbors.

```
MATCH   (B1:Building)<-[:Has_Proxi]->(B2:Building),
        (B1:Building)<-[:Has_Proxi]->(B3:Building)
RETURN B1
```

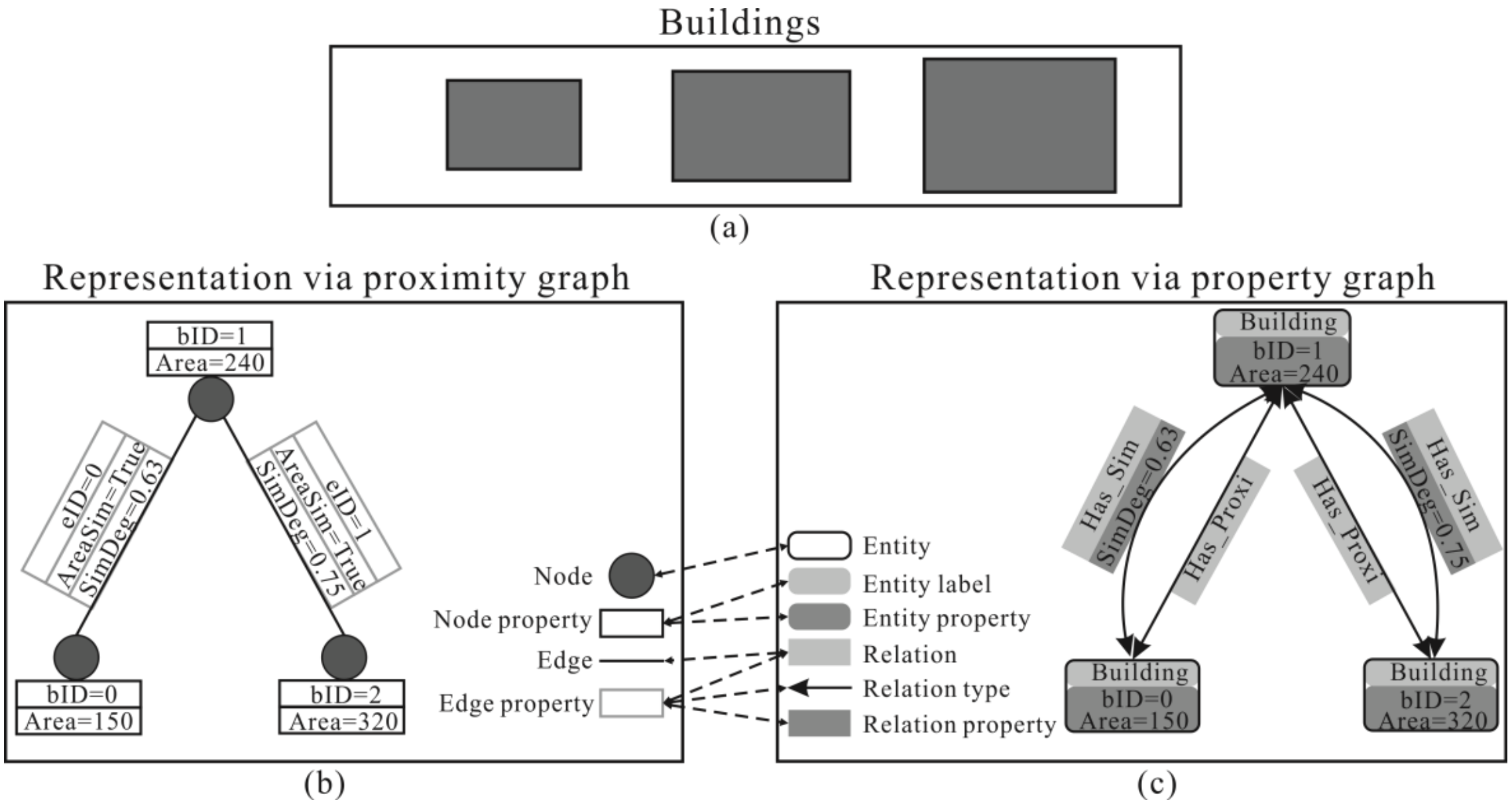

Figure 2. Representing the buildings and their relationships using proximity graph and property graph. (a) Buildings; (b) Representation via proximity graph; (c) Representation via property graph.

## 3. The knowledge graph representation for C-shaped building pattern recognition

As per the illustrations in related works, C-shaped buildings cannot be directly recognized through knowledge graphs. To recognize C-shaped building patterns via knowledge graphs, C-shaped buildings need to be recognized beforehand using existing methods. Therefore, the primary focus of our method is to recognize the C-shaped building arrangement initially and then expand the patterns at one scale based on the recognized patterns on other scales. Hence, it is crucial to represent the relations between buildings involving C-shaped building arrangement recognition at one scale and the corresponding relations between buildings across scales.

### 3.1 *Defining the building relations at one scale*

According to Du et al. (2015) and Gong et al. (2018), a C-shaped building arrangement comprises three adjacent buildings, with one building in the middle and two buildings on either side that form the wings of the C. The wing buildings are fully parallel to each other and partly perpendicular to the middle building. Therefore, to recognize a C-shaped building arrangement in a constructed knowledge graph, it is necessary to define the ‘proximity relation’, ‘fully parallel relation’, and ‘partly perpendicular relation’. And establishing 'interval relations' is the first step in defining the latter two.

(1) Proximity relation

The proximity relation between two buildings in topological maps is typically determined by adjacency or shared borders or corners. Because buildings separated by roads are not adjacent, we propose using the constrained Delaunay triangulation (CDT) skeleton that considers roads to detect the proximity relations between buildings according to Liu et al. (2014), as shown in Figure 3(a). Buildings that share the same CDT skeleton are considered proximate to each other, and a proximity graph $PG=(PE,PV)$ can then be built to represent the proximity relations between buildings, as shown in Figure 3(b). $PV=\{pv_1, pv_2, ..., pv_m\}$ is the node-set, where $pv_m$ represents building $B_m$. $pE=\{pe(pv_i, pv_j), pv_i \in PV, v_j \in PV\}$ is the edge set, where $pe(pv_i, pv_j)$ denotes that buildings $B_i$ and $B_j$ are proximate to each other. The proximity relation between $B_i$ and $B_j$, denoted as $P(B_i, B_j)$, is symmetric, which means $P(B_i, B_j) \leftrightarrow P(B_j, B_i)$.

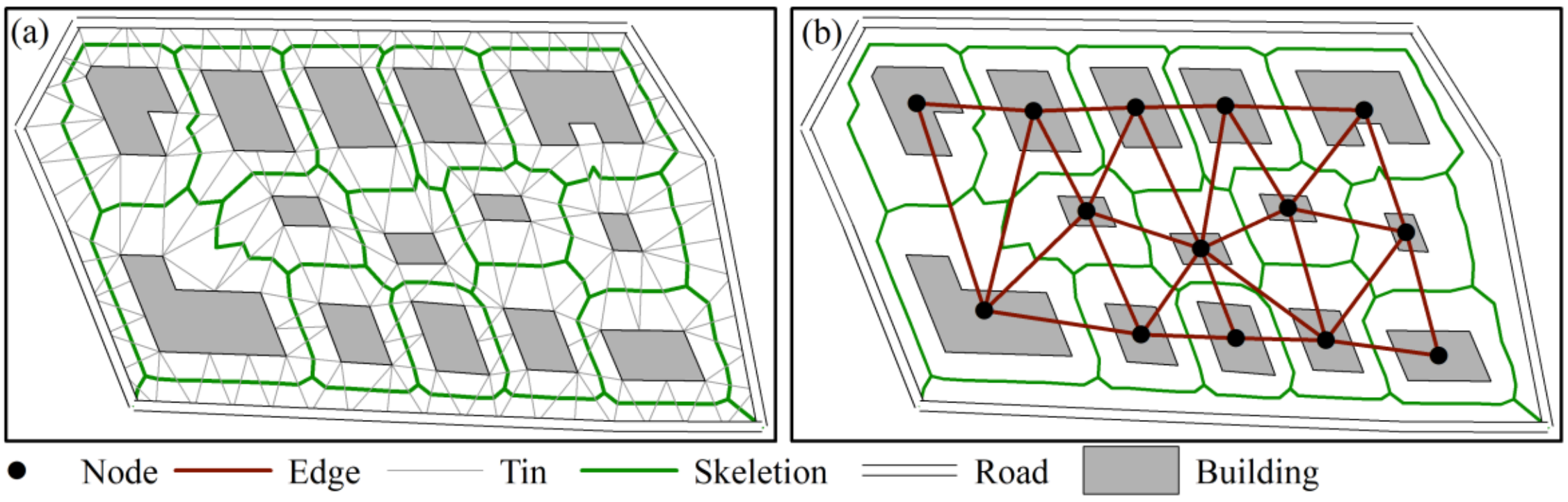

Figure 3. Proximity relations between buildings. (a) The constrained Delaunay triangulation (CDT) skeleton; (b) The proximity graph.

(2) Interval relation

We approximate the buildings by their smallest bounding rectangle (SBR) to define the interval relations according to Du et al., (2015). Suppose two building $B_a$ (the referent one) and $B_b$ (the target one), their SBRs denoted as SBR($B_a$) and SBR($B_b$), the $l_a$ and $s_a$ are long and short axes of SBR($B_a$), the two interval relations (Figure 4(a)) between $B_a$ and $B_b$ can be represented by projecting SBR($B_b$) to $l_a$ and $s_a$, denoted as $r(l_a, l_a^b)$ and $r(s_a, s_a^b)$, and form a relation $r(B_a, B_b) = r(l_a, l_a^b) \times r(s_a, s_a^b)$, where $l_a^b$ and $s_a^b$ are the projections. According to the interval algebra introduced by Allen (1983), $r(l_a, l_a^b)$ or $r(s_a, s_a^b)$ can be modeled as 13 basic relations, and $r(B_a, B_b)$ can then be defined as 169 basic relations, as shown in Figure 4(b). While non-overlapped relations are more prevalent in reality and can be abstracted into patterns, we do not consider overlapped interval relations (indicated by × in Figure 4). To prevent excessive over-approximation or overlapped interval relations, we incorporate the rectangularity ($S_{rec}$) to determine whether a building can be approximated by its SBR, where $S_{rec}$ is defined as Eq. (1).

$$S_{rec} = A / A_{sbr} \tag{1}$$

where $A$ is the area of the building, $A_{sbr}$ is the area of its SBR. If a building is with $S_{rec} \geq 0.6$, we approximate it using its SBR; else it indicates a complex shape that is unsuitable for pattern recognition, and we represent it as-is, without considering interval relations in our approach. To further measure the facing degree of two buildings ($B_a$ and $B_b$) in an interval relation, the facing ratio ($F_d$) is defined as Eq. (2).

$$F_d(B_a, B_b) = Max\{Overlap(l_a^b, l_a)/Merge(l_a^b, l_a), Overlap(s_a^b, s_a)/Merge(s_a^b, s_a)\} \quad (2)$$

where *Overlap* means an intersection, *Merge* means a union.

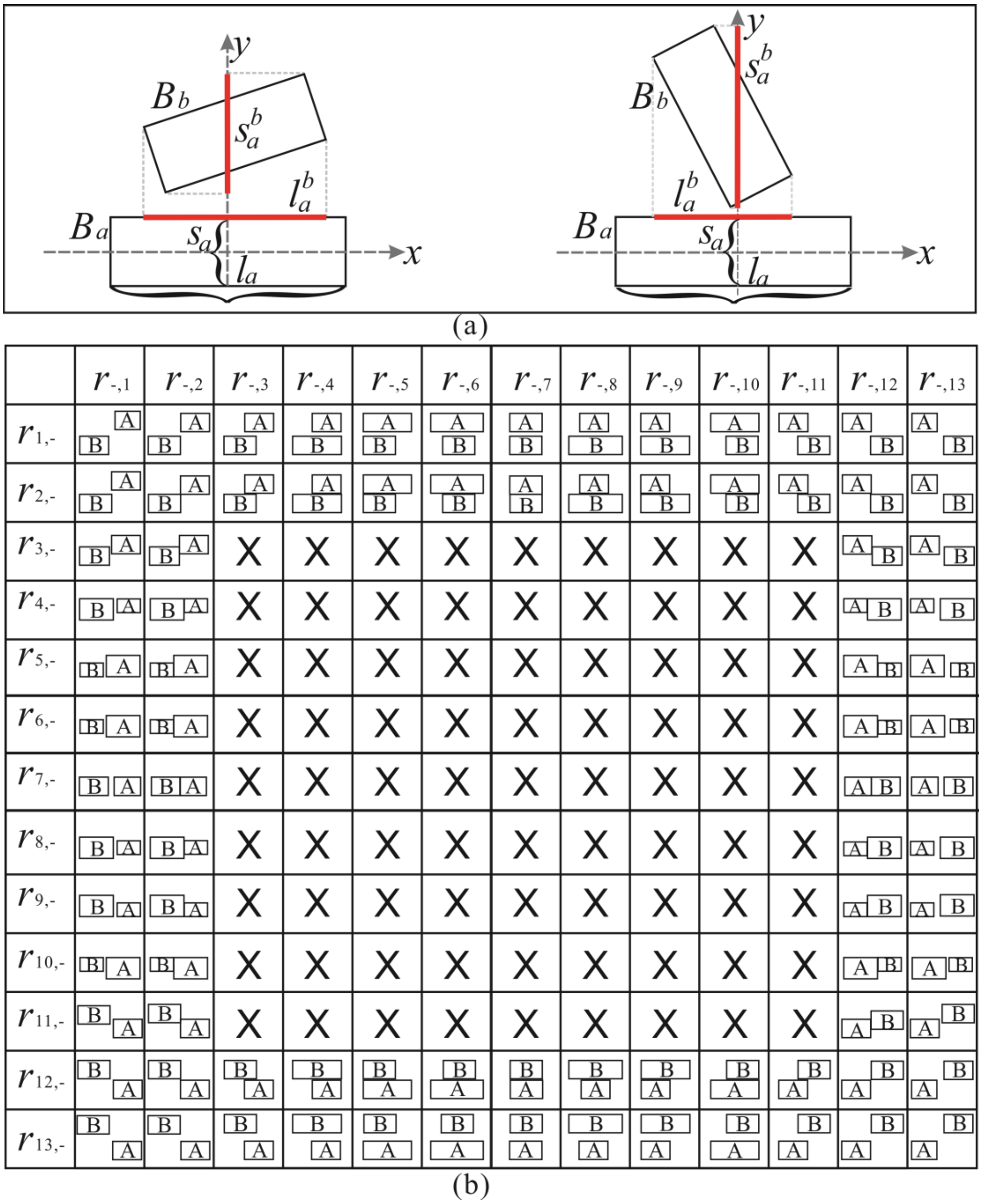

Figure 4. Interval relations definition between two buildings.

(3) Fully parallel relation

Two buildings $B_a$ and $B_b$ are fully parallel (*Full_Para*) if they have similar size

(*SimA*), parallel directions(*ParaO*), their long axes overlap over a certain degree and two buildings fully overlap on long axes only, and defined as Eq. (3) (Du et al., 2015).

$$\begin{aligned} &Full_Para(B_a, B_b) =_{def} \{SimA(B_a, B_b) \wedge ParaO(B_a, B_b) \wedge F_R(B_a, B_b) \geq \delta_1 \\ &\qquad \wedge r(B_a, B_b) \in \{r_{1,j}, r_{2,j}, r_{12,j}, r_{13,j}\}_{j=3}^{11}\} \\ &Where\ SimA(B_a, B_b) =_{def} (|A_a / A_b - 1| \leq \delta_2), \\ &\qquad ParaO(B_a, B_b) =_{def} (|O_a - O_b| \leq \delta_3 \vee (180 - |O_a - O_b|) \leq \delta_3) \end{aligned} \tag{3}$$

where $A$ is the building area, $O$ is the orientation of the SBR of a building, $\delta_1$, $\delta_2$, and $\delta_3$ are thresholds.

(4) Partly perpendicular relation

A buildings $B_b$ is partly perpendicular (*Part_Per*) to building $B_a$ if they are with perpendicular directions (*PerO*) and the projections of $B_b$ to $B_a$ either partly overlap with the axes of $B_a$, and is defined as Eq. (4).

$$\begin{aligned} &Part_Per(B_a, B_b) =_{def} \{PerO(B_a, B_b) \wedge r(B_a, B_b) \in \{r_{1,3}, r_{2,3}, r_{1,11}, r_{2,11}, r_{12,3}, r_{13,3}, r_{12,11}, r_{13,11}\}\} \\ &Where\ PerO(B_a, B_b) =_{def} \{(90 - |O_a - O_b|) \leq \delta_3\} \end{aligned} \tag{4}$$

where $O$ is the orientation of the SBR of a building.

### 3.2 *Defining the building relations across scales*

When buildings are represented at smaller scales, their boundaries are simplified and they may be eliminated, aggregated with others, or collapsed into points. This results in coarser or more abstract representations of buildings at larger scales. Based on the level of detail, six relations can occur between building features across different scales, as detailed in Table 2(Fan et al., 2014; Memduhoğlu & Basaraner, 2022). These relations include one-to-one, one-to-many, many-to-one, and many-to-many matches between corresponding buildings at two different scales. However, in some cases, a building at a scale may not have a corresponding match at the other scale, resulting in either a one-to-none or none-to-one relation.

Table 2. The definition for building relations across scales.

| |
|---|
| **One-to-one relation**: A building at one scale corresponds to one and only one at another scale; |
| **One-to-many relation**: A building at one scale corresponds to multiple buildings at another scale; |
| **Many-to-one relation**: Multiple buildings at one scale correspond to one building at another scale; |
| **Many-to-many relation**: Multiple buildings at one scale correspond to multiple buildings at another scale; |
| **One-to-none relation**: A building at one scale corresponds to no buildings at another scale; |
| **None-to-one relation**: No buildings at one scale correspond to any buildings at another scale. |

### 3.3 *Representing the building and their relations in knowledge graph*

The property graph represents the buildings and building relations within a knowledge graph as $G=(V,E)$, where $V=\{v_1,v_2,...,v_n\}$ refers to an entity set, $E=\{e(v_i,v_j),v_i \in V,v_j \in V\}$ is a relation set, $V$ and $E$ are defined as follows.

(1) Entities

The entities are individual buildings or building groups, each of which possesses a property as '*Scale*' that designates the scale of the associated building data. To differentiate between individual buildings and building groups, entities are assigned labels as '*SingleB*' and '*GroupB*'. As defined in Eqs. (3) and (4) for fully parallel relation (*Full_Para*) and partly perpendicular relation (*Part_Per*), area and orientation are involved. Therefore, the area and orientation are represented as the property of a building entity. Because we aim to recognize C-shaped patterns, a property '*ShapeC*' is given to indicate if an entity is a C-shaped one. Based on the above analysis, the entities in a knowledge graph are represented in Table 3.

Table 3. The entity representation in the knowledge graph.

| Class | Element | Description |
|---|---|---|
| Entity | $v_i$ | Indicate an entity |
| Entity label | *SingleB* | Indicate an entity as an individual building |
| | *Group B* | Indicate an entity as a building group |
| Entity property | *vID* | Unique identifier of an entity |

| | |
|---|---|
| *Scale* | Indicate the scale of the building data |
| *Area* | Indicate the area of a building |
| *Ori* | Indicate the orientation of the SBR of a building |
| *ShapeT* | Indicate whether an entity is a C-shaped one (*true* or *false*) |

(2) Relations

Recognizing C-shaped building patterns involves several types of relations at one scale, including proximity, interval, *Full_Para*, and *Part_Per*, as well as matching relations across scales. In addition, a '*Belong_To*' relation must be added to indicate that an individual building entity is part of a group building entity.

To represent relations at one scale, *Full_Para*, and *Part_Per* can be obtained based on the size, orientation, proximity, and interval relations between buildings. Therefore, we represent only proximity and interval relations in the knowledge graph, while *Full_Para* and *Part_Per* will be obtained based on rule-based reasoning in the built knowledge graph. To distinguish between the proximity and interval relations, we assigned the '*Has_Proxi*' and '*Has_Interval*' types to these relations. The basic type of an interval relation is denoted by a property as '*Inter_T*'. For instance, if buildings $B_i$ and $B_j$, represented as entity $v_i$ and $v_j$, are proximate to each other, a relation $e(v_i,v_j)$ is established with a type of '*Has_Proxi*', indicating that $e(v_i,v_j)$ is a proximity relation.

The relationships across scales between two buildings, including one-to-one, one-to-many, many-to-one, and many-to-many, can be represented as '*Has_Match*' with a property '*Match_T*', where '*Match_T*' values are one-to-one, one-to-many, many-to-one, and many-to-many. For instance, if two entities, $v_i$ and $v_j$, have a '*Has_Match*' relation with property '*Match_T*' of one-to-many, this implies a one-to-many relationship between $v_i$ and $v_j$. Additionally, none-to-one and one-to-none relationships can be explicitly expressed if they are not represented in a knowledge graph.

Based on the above analysis, Table 4 shows how the relations in a knowledge graph can be represented. Using this representation of entities and relations, we can construct and store a knowledge graph in Neo4j, an example of the constructed knowledge graph is shown in Figure 5. Note that different knowledge graphs can be built for this task, and we will discuss building a different knowledge graph in *Section* 6.1.

Table 4. The relation representation in the knowledge graph.

| **Class** | **Element** | **Description** |
|---|---|---|
| Relation | $e(v_i,v_j)$ | Indicates a relation |
| Relation type | *Has_Proxi* | Indicates that a relation is a proximity relation between two building entities |
| | *Has_Inter* | Indicates that a relation is an interval relation between two building entities |
| | *Has_Match* | Indicates that a relation is a matching relation between two entities |
| | *Belong_To* | Indicates a relation that a building entity is part of a building group entity |
| Relation property | *eID* | Unique identifier of a relation |
| | *Inter_T* | A property of relation '*Has_Inter*', indicates the type of '*Has_Inter*'. Suppose the type of a '*Has_Inter*' is $r_{i,j}$, the value of *Inter_T* is calculated as ($i$-1)*13+$j$ |
| | *Face_R* | A property of relation '*Has_Inter*', indicates the facing degree of two building s |
| | *Match_T* | A property of relation '*Has_Match*', indicates the type of '*Has_Match*' |

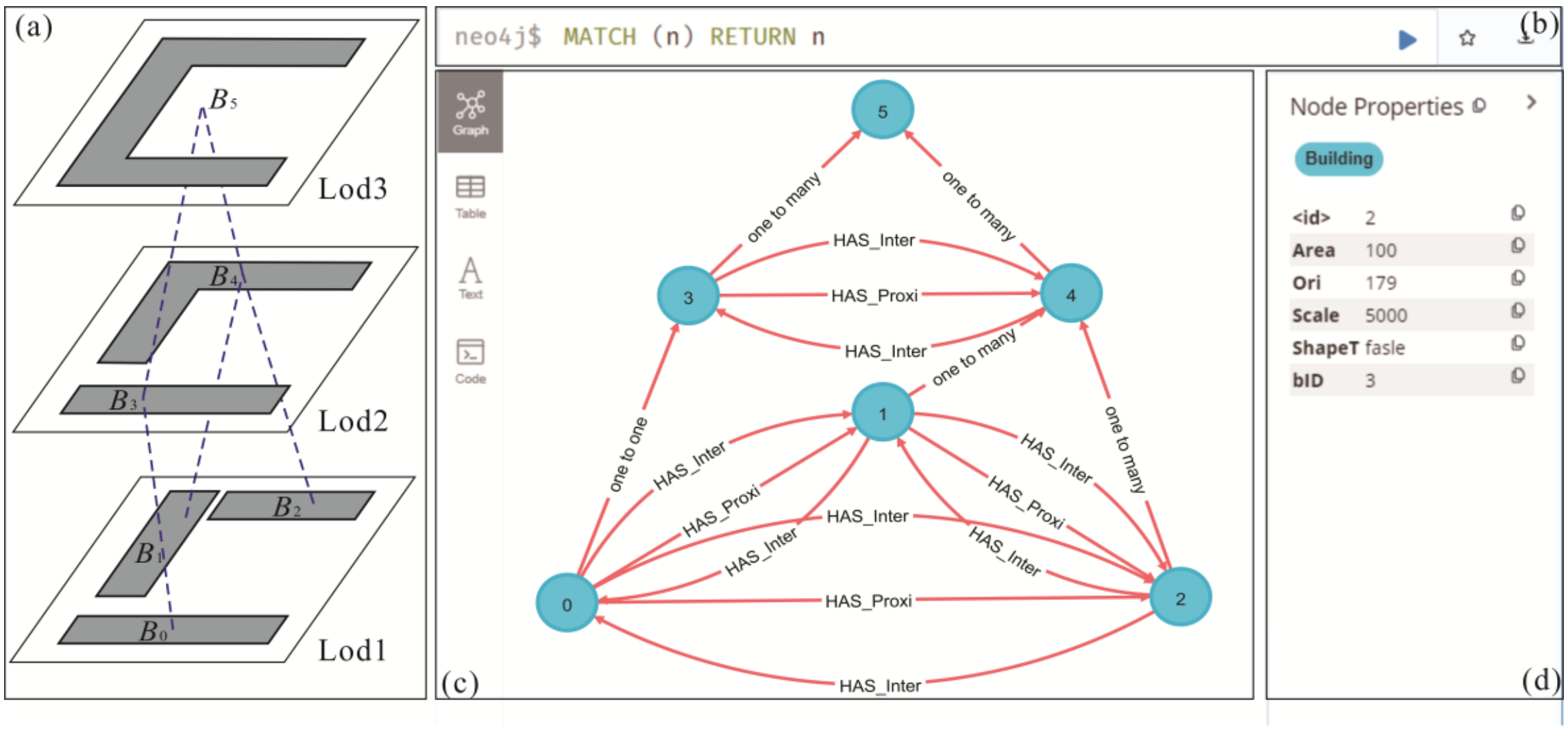

Figure 5. The built knowledge graph for multi-scale building data in Neo4j. (a) The multi-scale building data; (b) The query window; (c) The built knowledge graph; (d) The details for a selected entity or relation.

## 4. Rule-based reasoning for C-shaped building pattern recognition

### 4.1 *Recognize the C-shaped building arrangement at one scale*

A C-shaped building arrangement consists of three adjacent buildings, with one building in the middle ($B_M$) and two buildings on either side that form the wings of the C ($B_{W1}$ and $B_{W2}$). The wing buildings are fully parallel to each other, while also being partly perpendicular to the middle building, and a C-shaped building arrangement is defined as Eq. (5).

$$C_Pattern(B_M, B_{W1}, B_{W2}) =_{def} \{P(B_M, B_{W1}) \wedge P(B_M, B_{W2}) \wedge P(B_{W1}, B_{W2}) \\ \wedge Full_Para(B_{W1}, B_{W2}) \wedge Part_Per(B_M, B_{W1}) \wedge Part_Per(B_M, B_{W2})\} \quad (5)$$

To recognize the C-shaped building arrangement in a knowledge graph, the definition needs to be converted into rules that are suitable for rule-based reasoning in a knowledge graph. Since we store the knowledge graph in a Neo4j database, we use the Cypher language to define rules to derive the C-shaped building arrangement. As defined in Eq.(5), *Full_Para* and *Part_Per* need to be obtained because they are not represented in the knowledge graph. Additionally, similar size (*SimA*), parallel directions(*ParaO*), and perpendicular directions (*PerO*) need to be obtained first for *Full_Para* and *Part_Per* as defined in Eqs. (3) and (4). Therefore, the rule-based reasoning for the C-shaped building pattern consists of three steps: (1) verify the *SimA*, *ParaO*, and *PerO*; (2) verify the *Full-Para* and *Part_Per*; (3) recognize the C-shape building arrangements. The rules are presented in Table 5.

Table 5. Rules for C-shaped building recognition via Cypher language in the built knowledge graph.

```
MATCH (B1:SingleB)<-[R1:Has_Proxi]-> (B2:SingleB)
WHERE abs(B1.Area/B2.Area-1) < δ2
        MERGE (B1)-[r:SimA]-(B2)
WHERE abs(B1.Ori-B2.Ori) ≤ δ3 OR abs(180-(B1.Ori-B2.Ori)) ≤ δ3
        MERGE (B1)-[r: ParaO]-(B2)
WHERE abs(abs(B1.Ori-B2.Ori)-90) ≤ δ4
        MERGE (B1)-[r: PerO]-(B2)

MATCH (B1:SingleB)-[R1:SimA] -(B2:SingleB),
        (B1:SingleB)-[R2: ParaO] -(B2:SingleB)
        (B1:SingleB)-[R3:Has_Interval] ->(B2:SingleB)
WHERE Round(R3.Inter_T/13,0) IN [1, 2, 12, 13] AND R3.Inter_T%13 IN [3, 4, 5, 6, 7, 8, 9,
10,11] AND R3.Face_R ≥ δ4
        Create (B1)-[r:Full_Para]->(B2)
WHERE R3.Inter_T IN  [3, 16, 11, 24, 146,159,154,167]
        Create (B1)-[r:Part_Per]->(B2)

MATCH (B1:SingleB)-[: Full_Pal] ->(B2:SingleB),
        (B1:SingleB)-[: PerO]->(B3:SingleB)
        (B2:SingleB)-[: PerO]->(B3:SingleB)
WITH B1, B2, B3
        MERGE (BG1: GroupB{Shape_T: true, Scale:B1.Scale})
        MERGE (B1)-[:BelongsTo]->(BG1)
        MERGE (B2)-[:BelongsTo]->(BG1)
        MERGE (B3)-[:BelongsTo]->(BG1)
```

### 4.2 *Enrich the structural building patterns across scales*

Our method considers two types of enrichments, as identified by Wei et al. (2022) and Xing et al. (2021). Wei et al. (2022) propose decomposing a building into sub-buildings for pattern recognition. Buildings at larger scales represent potential subdivisions of the buildings at a smaller scale, allowing structural patterns recognized at a larger scale to enrich those recognized at a smaller scale (i.e., **bottom-up enrichment**). In contrast, Xing et al. (2021) suggest grouping buildings into clusters for pattern recognition. Buildings at smaller scales represent potential building groups at a larger scale, enabling structural patterns recognized at a smaller scale to enrich those recognized at a larger scale (i.e., **up-bottom enrichment**).

4.2.1 *Bottom-up enrichment*

For bottom-up enrichment, suppose three buildings ($B_a$, $B_b$, $B_c$) as a group are recognized at a larger scale ($S_b$) as a C-shaped pattern. Their corresponding buildings at a smaller scale ($S_u$) are $BS=\{B_i\}_{i=1}^{n}(n\geq 1)$. As proposed by Wei et al. (2022), a building at a smaller scale can be decomposed into sub-buildings for building pattern recognition. If $B_a$, $B_b$, $B_c$ are part of $B_i$ (i.e, have a many-to-one or one-to-one relation), then $BS=\{B_i\}_{i=1}^{n}(n\geq 1)$ can potentially form a C-shaped pattern. Take the C-shaped building pattern ($B_0$, $B_1$, $B_2$) in Figure 5(a) as an example, $B_3$ at a larger scale has a one-to-one relation with $B_0$, $B_4$ at a larger scale has a one-to-many relation to $B_1$, and $B_2$, then $B_3$ and $B_4$ can potentially form a C-shaped pattern building pattern. The rules for bottom-up enrichment via Cypher language in the built knowledge graph are defined in Table 6.

Table 6. Rules for bottom-up enrichment via Cypher language in the built knowledge graph.

```
MATCH (BG1:GroupB{Shape_T: true, Scale:Sb})-[:Belong_to]->(B1:SingleB)
WITH collect(B1) AS SB1
MATCH (B1)-[:Has_Match {Match_T: IN [one-to-one, one-to-many]]->(B2:SingleB {Scale: Su})
WITH collect(B2) AS SB2
MATCH (B2)-[:Has_Match {Match_T: IN [one-to-one, one-to-many]]->(B3:SingleB {Scale:Sb })
WITH collect(B3) AS SB3
WHERE SB1<>SB3
    MERGE (SB2: GroupB {Shape_T: true, Scale: Su})
```

4.2.2 *Up-bottom enrichment*

For up-bottom enrichment, suppose a building or a building group is recognized as a C-shaped pattern at a smaller scale ($S_u$). Their corresponding building groups at a larger scale ($S_b$) are $BS = \{B_i\}_{i=1}^{n} (n \geq 1)$. As proposed by Xing et al. (2021), buildings can be grouped into clusters for pattern recognition. If $BS = \{B_i\}_{i=1}^{n} (n \geq 1)$ are parts of $B_a$, (i.e., have a one-to-many or one-to-one relation), then $BS = \{B_i\}_{i=1}^{n} (n \geq 1)$ can potentially form a structural pattern. Take the C-shaped building pattern $B_5$ in Figure 5(a) as an example, $B_3$ and $B_4$ at a larger scale have a one-to-many relation to $B_5$, then $B_3$ and $B_4$ can potentially form a C-shaped pattern building pattern. Therefore, the Bottom-up enrichment can be implemented via Cypher language in the built knowledge graph, as defined in Table 7.

Table 7. Rules for bottom-up enrichment via Cypher language in the built knowledge graph.

| |
|---|
| MATCH (BG1:GroupB{Shape_T: true, Scale: $S_u$})-[:Belong_to]->(B1:SingleB)<br>WITH collect(B1) AS SB1<br>MATCH (B1)-[:Has_Match {Match_T: IN [one-to-one, one-to-many]]->(B2:SingleB {Scale: $S_b$})<br>WITH collect(B2) AS SB2<br>MATCH (B2)-[:Has_Match {Match_T: IN [one-to-one, one-to-many]]->(B3:SingleB {Scale: $S_u$})<br>WITH collect(B3) AS SB3<br>WHERE SB1<>SB3<br>MERGE (SB2: GroupB {Shape_T: true, Scale: $S_b$}) |
| MATCH (BG1:SingleB{Shape_T:true, Scale: $S_u$ })-[:Has_Match]->(B1:SingleB {Scale: $S_b$})<br>WITH collect(B1) AS SB1<br>MATCH (B1)-[:Has_Match {Match_T: IN [one-to-one, one-to-many]]->(B2:SingleB{Scale: $S_u$})<br>WITH collect(B2) AS SB2<br>WHERE SB1<>SB2<br>MERGE (SB1: GroupB, {Shape_T: true, Scale: $S_b$}) |

## 5. Experiments

### 5.1 *Experimental data and environment*

#### 5.1.1 *Experimental data*

The study area comprised 85,820 buildings located in Nanjing, China, which were collected from the Gaode Map. The area was viewed at a nominal scale of level 13, with a scale of approximately 1:5000, as per the levels of detail defined in the Gaode Map. A subset of the dataset, consisting of 1977 buildings, was manually generalized into two coarser levels (levels 11 and 12), based on the Gaode Map's level of detail, the resulting dataset is shown in Figure 6. The generalized strategies used for the dataset are presented in Table 8.

Table 8. The generalized strategies used for the dataset.

| Data | Represent features | Generalized strategies |
|---|---|---|
| Lod1 | Building with floor details(Level 13) | As-is |
| Lod2 | Building outlines (Level 12) | Simplification, delete small buildings, aggregate touched buildings |
| Lod3 | Building outlines or built-up areas (Level 11) | Simplification, delete small buildings, aggregate nearby buildings, typification for low-density area |

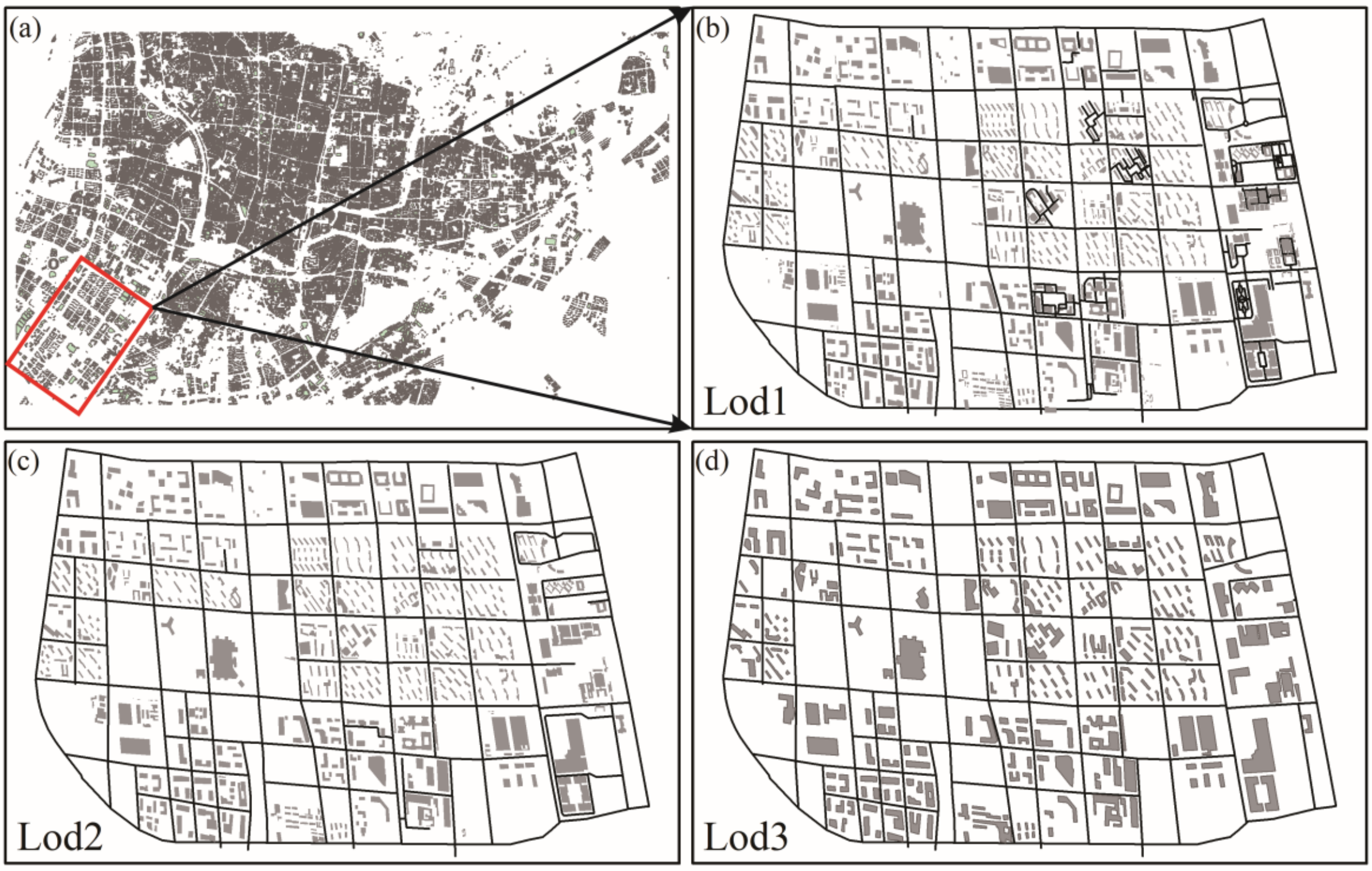

Figure 6. Experiment data.

Statistical analysis was conducted to evaluate the spatial characteristics of buildings, which included building count ($B_c$), average area ($Ave_A$), average edge count ($Ave_E_d$), the rate of buildings with edge count fewer than eight ($R_{Ed \leq 8}$), average rectangularity ($Ave_S_{rec}$), and the rate of buildings with rectangularity greater than 0.6 ($R_{Srec \geq 0.6}$) according to Wei et al. (2022). These parameters were computed for all buildings in the experimental data and are presented in Table 9. From the results in Table 9, we can make the following observations. Firstly, the majority of the buildings have simple shapes, as indicated by the high values of $R_{Ed \leq 8}$ (>0.64) and $R_{Srec \geq 0.6}$ (>0.78), This implies that the buildings can be approximated with their SBR for pattern recognition purposes. Secondly, as the scale of the buildings increases, the *Ave_Area* increases, while $R_{Ed \leq 8}$ , *Ave*_$S_{rec}$, and $R_{Srec \geq 0.6}$ decrease. This suggests that the buildings become more complex in shape and less likely to be accurately approximated with their SBR for pattern recognition. Therefore, shape recognition can be a crucial factor for enriching pattern recognition as the scale of buildings increases.

Table 9. Analysis of the spatial characteristics of buildings in the study area.

| Data | $B_c$ | *Ave_Area*/m$^2$ | *Ave*_$E_d$ | $R_{Ed \leq 8}$ | *Ave*_$S_{rec}$ | $R_{Srec \geq 0.6}$ |
|---|---|---|---|---|---|---|
| Lod1 | 1977 | 797.22 | 8.78 | 0.69 | 0.80 | 0.89 |
| Lod2 | 618 | 2528.14 | 9.48 | 0.67 | 0.80 | 0.83 |
| Lod3 | 337 | 4602.54 | 8.18 | 0.64 | 0.78 | 0.78 |

5.1.2 *Experimental environment*

We utilized the C# code to compute the measures for spatial characteristics of buildings and their relationships based on ArcEngine 10.2. The resulting knowledge graph was stored in Neo4j 1.4.12, and rule-based reasoning was implemented using Cypher language. The experiments were conducted on a personal computer equipped with an Intel® Core™ 1.60 GHz i5-8265U CPU and 8G RAM.

## 5.2 *Built knowledge graph and visualization*

(1) Deriving the building relations at one scale

*Section* 3.1 outlines the derivation process of building relations at one scale. Firstly, a proximity graph was proposed to represent proximity relations between buildings, as defined in *Section* 3.1. Based on the proximity relations, we derived the interval relations, fully parallel relations (*Full_Para*), and partly perpendicular relations (*Part_Per*) between buildings. To define *Full_Para* and *Part_Per* relations, we used the thresholds according to Du et al. (2015) and Wei et al. (2022) in Eqs. (3) and (4) as follows: $\delta_1$ was the threshold for determining the facing degree between two buildings to establish a *Full_Para* and was set to a value of 0.4; The threshold $\delta_2$ was used to assess the similarity in size between buildings and was set to a value of 2; The threshold $\delta_3$ was utilized to define the parallel or perpendicular orientation between two buildings, and was set to a value of 15.

(2) Deriving the building relations for across scale

To identify the corresponding relations between buildings at different scales, a geometric matching process can be performed. However, since geometric matching is not the primary focus of our approach, we employed a simple but efficient method proposed by Fan et al. (2014) and Memduhoglu et al. (2022) and identified corresponding buildings ($B_{ref}$ and $B_{tar}$) based on their overlapping areas, as defined in Eq. (6). Specifically, if $R_{overlap} \geq 0.3$, $B_{ref}$ and $B_{tar}$ were considered to be corresponding buildings. To ensure more accurate identification of corresponding relations, we manually adjusted the corresponding relations between buildings after the above processing.

$$R_{overlap} = \frac{A_{ref \cap tar}}{Min(A_{ref}, A_{tar})} \quad (6)$$

where $A_{ref}$ and $A_{tar}$ are the areas of $B_{ref}$ and $B_{tar}$, and $A_{ref \cap tar}$ is the area of the overlapping area of $B_{ref}$ and $B_{tar}$.

(3) The knowledge graph and visualization

Building upon the illustrations presented in *Section* 3.3, we can represent the inter-building relations within and across different scales as a knowledge graph, as depicted in Figure 7. Figure 7(a) visualizes the proximity relation at one scale (Has_Proxi) and the corresponding relations across scales (Has_Match), which was implemented via EChart, and a force-direct method was applied to avoid crosses and overlaps between nodes and edges as much as possible (Li et al., 2018). Figure 7(a) demonstrates that the defined proximity relation results in the formation of a cluster of building entities. Moreover, Figure 7(b) provides a more detailed view of the constructed knowledge graph, effectively showcasing how the inter-building relationships in a local area can be accurately represented as a knowledge graph.

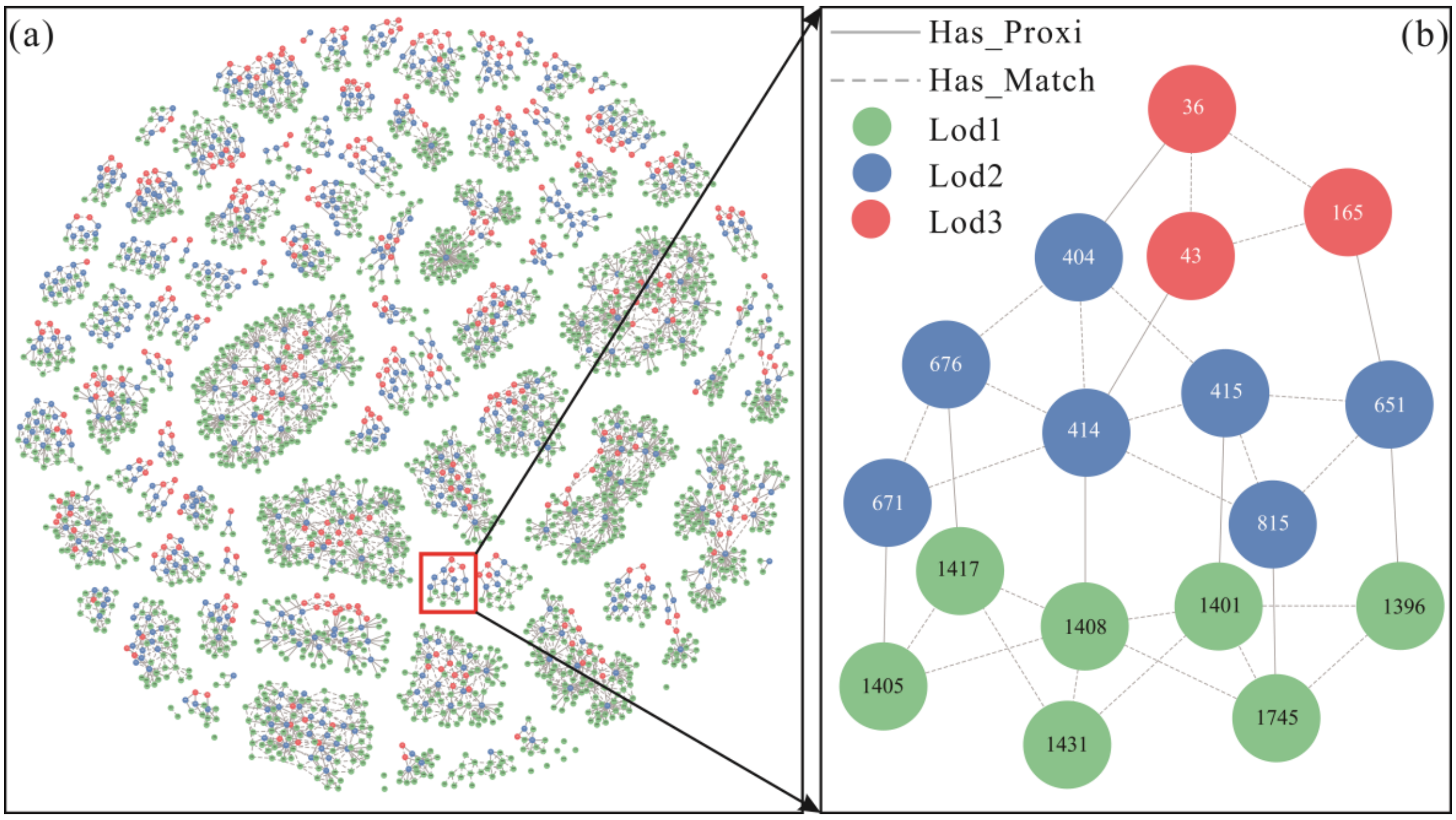

Figure 7. Built Knowledge graph.

### 5.3 *Pattern recognition results*

#### 5.2.1 *Effectiveness evaluation*

To evaluate the effectiveness of our approach, we conducted a comparative analysis with results identified by three graduate students experienced in cartography and geographic information science. The students independently annotated the C-shaped patterns with informed them that the buildings could be grouped and decomposed for pattern recognition. In cases where annotated results were inconsistent, a voting system was implemented. Recognized patterns that were consistent with those identified by the students were denoted as true positives (*tp*), those that were inconsistent as false positives (*fp*), and patterns not recognized by the proposed method as false negatives (*fn*). *Precision* and *Recall* values were calculated using equations (7) and (8) as defined in Du et al. (2015).

$$Precision = tp/(tp + fp) \tag{7}$$

$$Recall = tp/(tp + fn) \tag{8}$$

The comparative analysis was carried out in two parts: (1) recognition of C-shaped buildings and arrangements without enrichment (*SP+PP*), and (2) recognition of C-shaped buildings and arrangements with enrichment (*SP+PP+EP*). C-shaped buildings were recognized through manual shape recognition and C-shaped arrangements were recognized using the defined rules in *Section* 4.1. The pattern recognition results via the proposed method are presented in Figure 8, and the statistics on the comparison are presented in Table 10.

Table 10. Evaluation of the effectiveness.

| Level | Result | *tp* | *fp* | *fn* | *Precision* | *Recall* |
|---|---|---|---|---|---|---|
| Lod 1 | *SP+PP* | 23 | 4 | 11 | 85.2% | 67.7% |
| | *SP+PP+EP* | 32 | 6 | 2 | 84.2% | 94.1% |
| Lod 2 | *SP+PP* | 31 | 0 | 9 | 100% | 77.5% |
| | *SP+PP+EP* | 39 | 3 | 1 | 92.9% | 97.5% |
| Lod 3 | *SP+PP* | 27 | 0 | 6 | 100% | 81.8% |
| | *SP+PP+EP* | 30 | 2 | 3 | 97.0% | 90.9% |

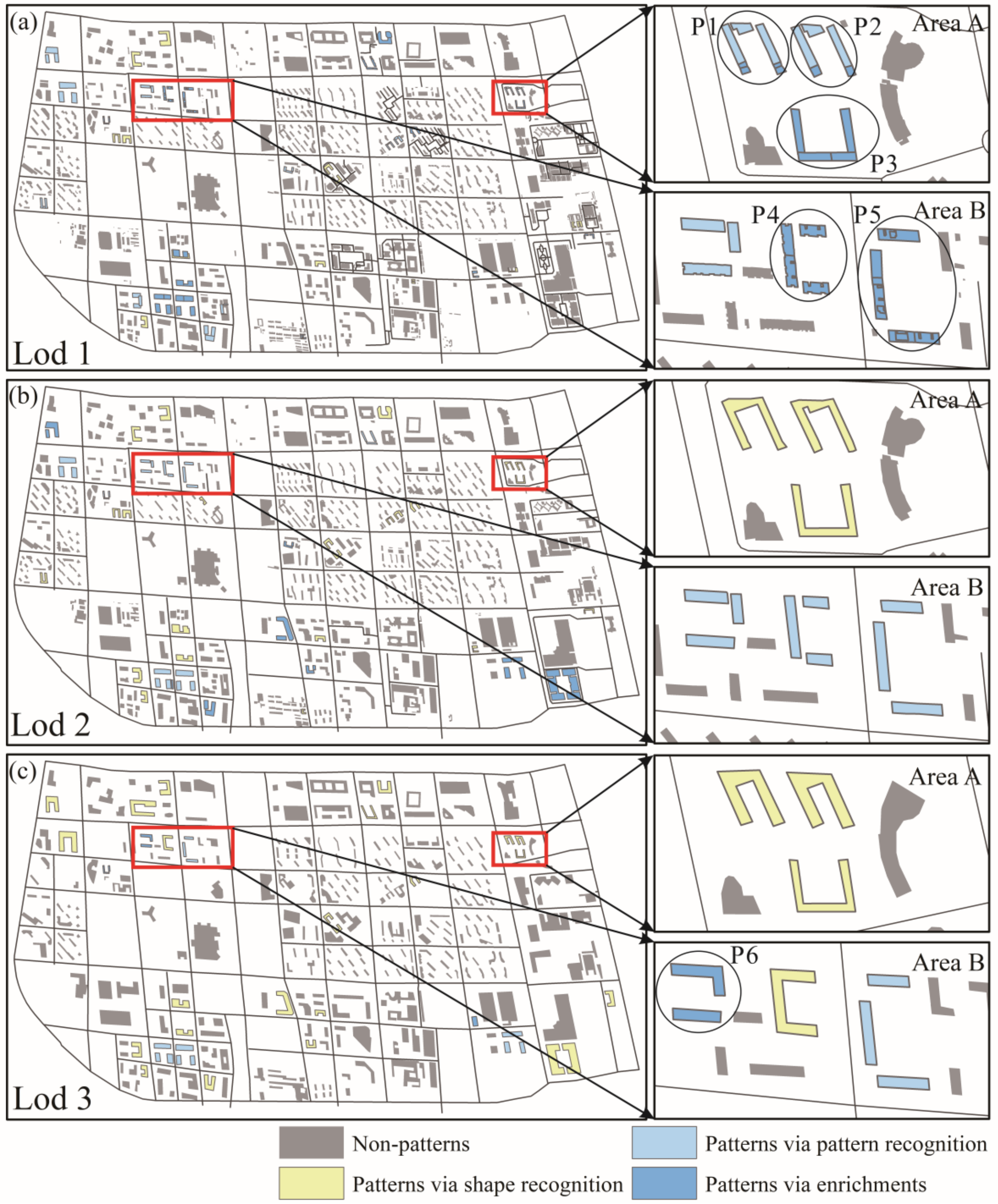

Figure 8. C-shaped building pattern recognition results.

In Figure 8 and Table 10, we observe that enrichments are necessary to recognize most C-shaped building patterns in the datasets. The *Recall* for Lod1 is 94.1%, for Lod2 is 97.5%, and for Lod3 is 90.9%, which is higher than the results without enrichments, which had a *Recall* of 26.4% for Lod1, 20.0% for Lod2, and 9.1% for Lod3. This difference can be attributed to the fact that human vision is a part-based system, and object-independent pattern recognition is insufficient to recognize all potential patterns. For example, the C-shaped arrangements P1 and P2 (Area A in Figure 9) can only be partially recognized, and P3, P4, P5, and P6 can’t be recognized with the existing method (Area A and Area B in Figure 9). However, all of these patterns can be recognized using the proposed approach, and these patterns are also visually aware patterns by users. Particularly, P1, P2, P3, P4, and P5 are recognized with up-down enrichments, and P6 is recognized with bottom-up enrichments. These comparisons demonstrate that enrichments can help recognize more building patterns across scales, as part-whole hierarchies can enhance the semantic concepts and relations of map data to a greater extent.

As for the *Precision* in Figure 8 and Table 10, we observe that high *Precision* can be achieved if enrichments are implemented, with a *Precision* of 84.2% for Lod1, 92.9% for Lod2, and 97% for Lod3. However, implementing enrichments can also decrease *Precision*, with a *Precision* decrease of 1.0% for Lod1, 7.1% for Lod2, and 3% for Lod3. These indicate that although enrichments can help recognize more patterns, they can also misrecognize some patterns, such as those shown in Figure 9. This occurs because patterns are only stable within a certain scale range, and as the scale changes, patterns at different scales may change (Tenerelli et al., 2011). For example, the C-shaped building in 9(d) can be recognized via shape recognition, but its corresponding building at higher levels is more likely to be rectangular, which will lead to a misrecognition in bottom-up enrichment. Therefore, our method may be more efficient for multi-scale data with a smaller-scale variation. To improve the effectiveness of pattern recognition across scales, a study on the variation law of building patterns with scale may be necessary.

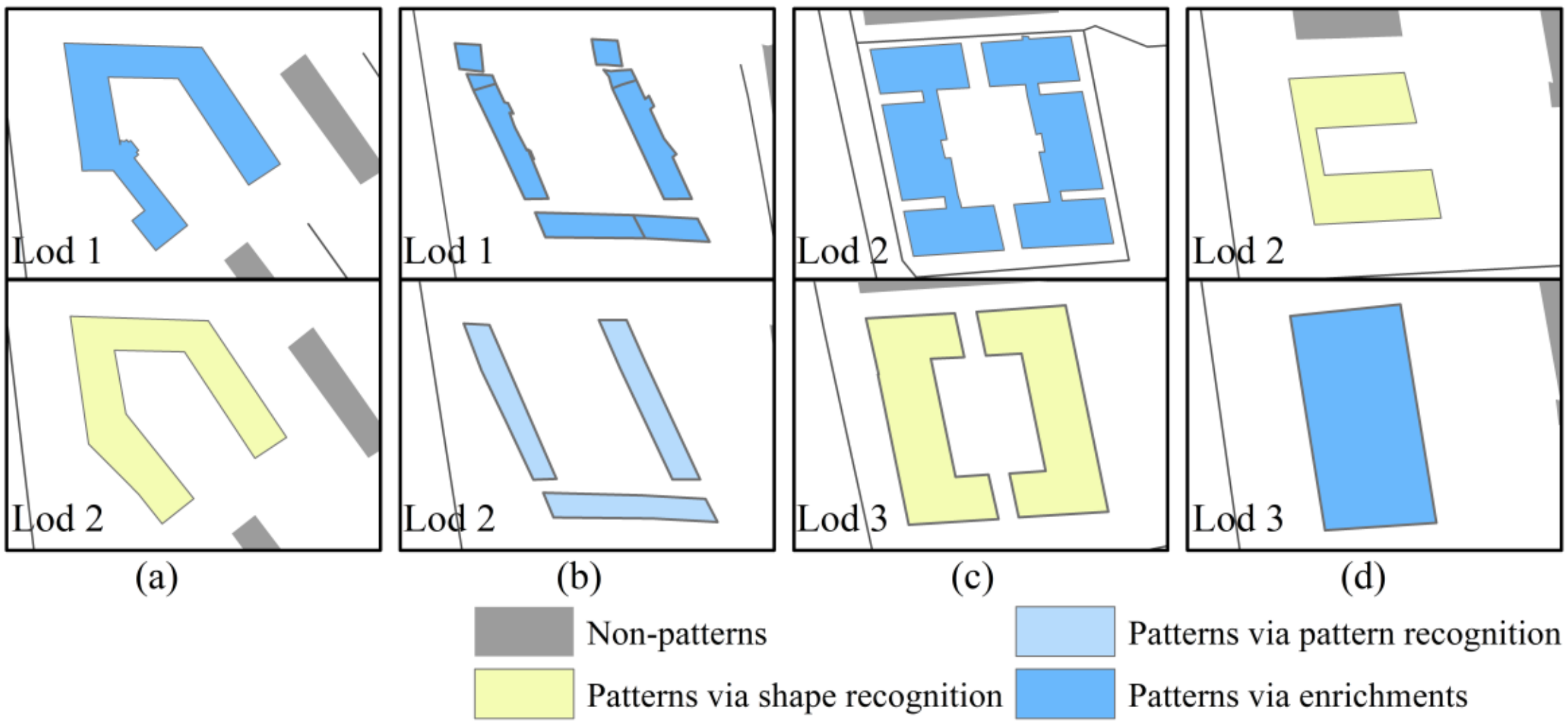

Figure 9. C-shaped building pattern misrecognition results.

5.2.2 *Efficiency evaluation*

To assess the efficiency of our approach, we measured the rule-based reasoning time for C-shaped building pattern recognition in the built knowledge graph. The proposed method was independently tested in 10 experiments, and the entity count ($v_c$), minimum time (*Min_t*), maximum time (*Max_t*), and average time (*Ave_t*) are presented in Table 11. As shown in the table, the *Ave_t* for Lod1 is 1.209s, for Lod2 is 0.673s, and for Lod3 is 0.431s, indicating good efficiency. Moreover, the *Ave_t* exhibits an approximately linear increase with the increase in $v_c$, suggesting that the *Ave_t* won't sharply increase with larger data sizes, making it suitable for large areas. This is because the knowledge graph stores the relations as key-value pairs in a graph database, eliminating the need for computing resources to establish matching between entities, resulting in efficient reasoning for new knowledge.

On the other hand, existing methods for recognizing C-shaped building arrangements also use rule reasoning, but they mostly use the proximity graph instead of the knowledge graph employed in our proposed approach. To compare the efficiency of these methods, we also recorded the average time (*Ave_t*) required to recognize the C-shaped arrangements using the defined rules in Table 5 via the proximity graph. The *Ave_t* for Lod1 was 0.321s, Lod2 was 0.765s, and Lod3 was 12.787s. Compared to the proposed approach, the speeds are 0.91, 1.37, and 9.35 times faster, respectively. Moreover, our method's superior performance was even more pronounced when processing larger data sets. Therefore, we can conclude that our method is more efficient and provides significant efficiency improvement as data size increases.

Table 11. Evaluation of the efficiency.

| Data | $v_c$ | *Min_t/s* | *Max_t/s* | *Ave_t/s* |
|---|---|---|---|---|
| Lod1 | 1977 | 0.841 | 1.677 | 1.368 |
| Lod2 | 618 | 0.420 | 0.902 | 0.558 |
| Lod3 | 337 | 0.324 | 0.468 | 0.351 |

## 6. Discussion

### 6.1 *Building a different knowledge graph*

The extensibility of the knowledge graph is a key advantage that we leverage to construct the knowledge map in a flexible and scalable bottom-up manner, tailored to the specific application requirements. In practice, different knowledge graphs can be constructed based on diverse needs. For instance, we have represented building size, orientation, proximity, and interval relations in the knowledge graph, while *Full_Para* and *Part_Per* are obtained through rule-based reasoning. Alternatively, we can directly represent *Full_Para* and *Part_Per* to build a different knowledge graph, as shown in Table 12. This would also simplify the rules for pattern recognition, as illustrated in Table 13. However, obtaining *Full_Para* and *Part_Per* before building the knowledge graph would take a longer time. Despite this, it would also enable recognition of the C-shaped building patterns.

Table 12. The elements of knowledge graph for C-shaped building pattern recognition.

| **Class** | **Element** | **Description** |
|---|---|---|
| Entity | $v_i$ | Indicate an entity |
| Entity label | *SingleB* | Indicate an entity as an individual building |
| | *Group B* | Indicate an entity as a building group |
| Entity property | *vID* | Unique identifier of an entity |
| | *Scale* | Indicate the scale of the building data |
| | *ShapeT* | Indicate whether an entity is a C-shaped one (*true* or *false*) |
| Relation | $e(v_i,v_j)$ | Indicates a relation |
| Relation type | *Full_Para* | Indicates that a relation is a matching relation between two entities |
| | *Part_Per* | Indicates that a relation is a matching relation between two entities |
| | *Has_Match* | Indicates that a relation is a matching relation between two entities |
| | *Belong_To* | Indicates a relation that a building entity is part of a building group entity |
| Relation property | *eID* | Unique identifier of a relation |
| | *Match_T* | A property of relation '*Has_Match*', indicates the type of '*Has_Match*' |

Table 13.The rules for C-shaped building pattern recognition.

```
MATCH (B1:SingleB)-[: Full_Pal] ->(B2:SingleB),
        (B1:SingleB)-[: PerO]->(B3:SingleB)
        (B2:SingleB)-[: PerO]->(B3:SingleB)
WITH B1, B2, B3
       MERGE (BG1: GroupB {Shape_T: true, Scale: B1.Scale})
       MERGE (B1)-[:BelongsTo]->(BG1)
       MERGE (B2)-[:BelongsTo]->(BG1)
       MERGE (B3)-[:BelongsTo]->(BG1)
```

## 6.2 *Recognizing other kinds of structural patterns*

To facilitate the recognition of C-shaped building patterns, we represent the underlying relations as a knowledge graph and formulate the rules for pattern recognition and enrichment as query conditions on the graph. By extending this method to other building patterns, such as the example E-shaped or H-shaped patterns depicted in Figure 10, we can achieve similar recognition and enrichment results. It is worth noting that certain patterns, such as patterns colored light lavender in Figure 10(a), require cross-scale enrichment for recognition, which cannot be achieved using existing approaches.

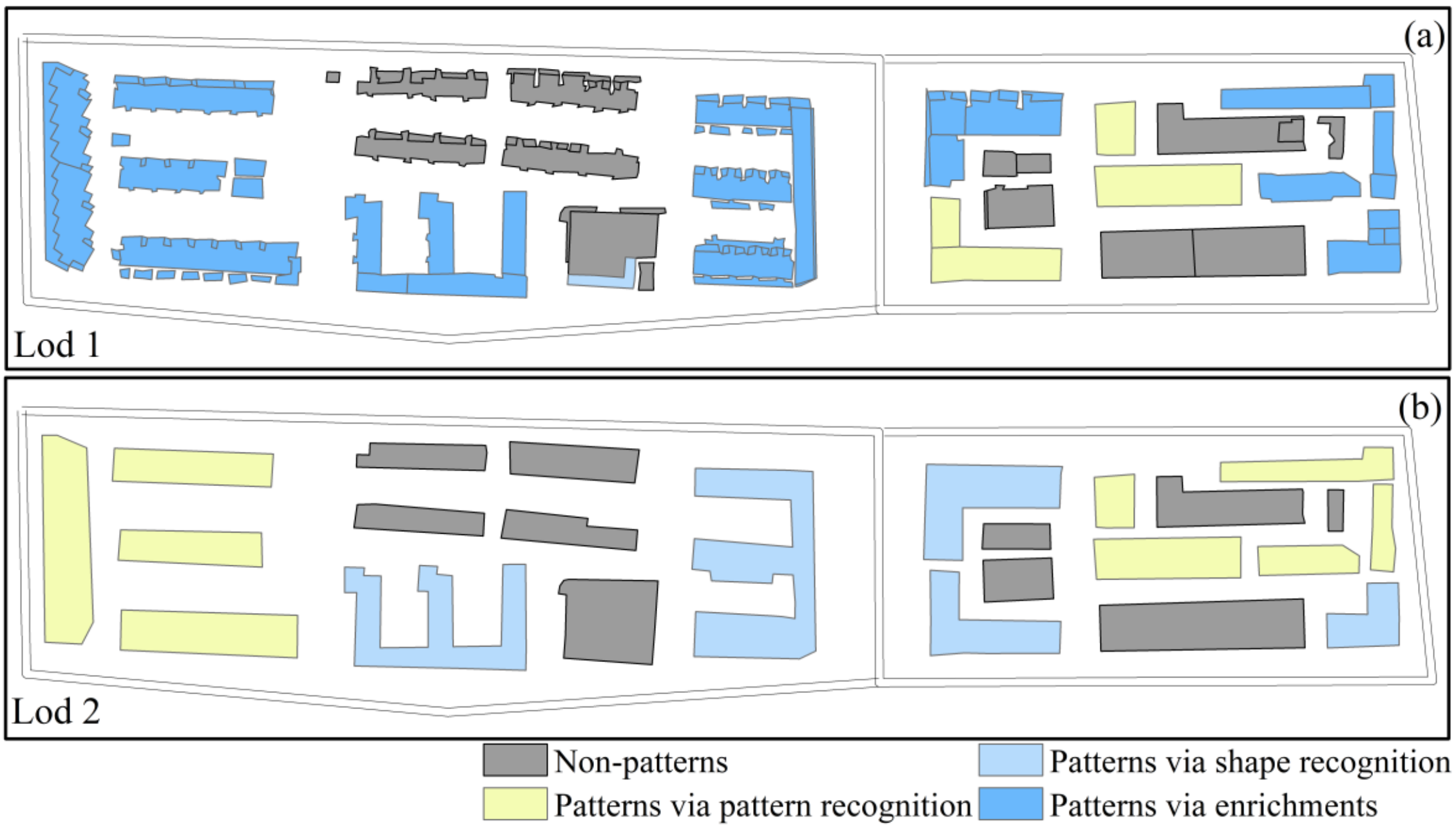

Figure 10. Recognition of other kinds of structural patterns.

## 7. Conclusion

To improve the efficiency and effectiveness of recognizing building patterns in topographical maps, we propose a method that integrates multi-scale data using a knowledge graph. As a case study, we focused on the C-shaped building pattern and represented the relationships between buildings within and across different scales in a knowledge graph. We then used rule-based reasoning to recognize and enrich patterns across scales within the built knowledge graph. Our results demonstrated that our method achieved higher precision and efficiency compared to existing approaches, with a recall of 26.4% for LOD1, 20.0% for LOD2, and 9.1% for LOD3, and an efficiency improvement of 0.91, 1.37, and 9.35 times. Additionally, our proposed approach was capable of recognizing other building patterns, such as E-shaped and H-shaped patterns.

However, we also found that patterns were only stable within a certain range of scales. As the scale changed, the patterns at different scales could change, leading to misrecognition. Therefore, future work will focus on understanding the variation law of building patterns with scale to enable more effective pattern recognition across scales.

## Disclosure statement

No potential conflict of interest was reported by the author(s).

## Data and code availability statement and data deposition

The data and code that support the findings of this study are all openly available, website is: https://github.com/TrentonWei/Linear-Building-Pattern-Recognition-.git

## Acknowledgements

This work was supported by The National Natural Science Foundation of China (grant agreement No 41871378) and The Youth Innovation Promotion Association Foundation of Chinese Academic of Sciences(grant agreement No Y9C0060).

The authors would like to thank Dr. Yuangang Liu, the editors, and the anonymous reviewers for their helpful and constructive comments that greatly contributed to improving the manuscript.